\documentclass{article}

\usepackage{arxiv}

\usepackage[utf8]{inputenc} 
\usepackage[T1]{fontenc}    
\usepackage{hyperref}       
\usepackage{url}            
\usepackage{booktabs}       
\usepackage{amsfonts}       
\usepackage{nicefrac}       
\usepackage{microtype}      
\usepackage{lipsum}
\usepackage{graphicx}
\usepackage{subfig}
\usepackage{float}
\usepackage{placeins}
\usepackage{amsmath}
\usepackage[dvipsnames]{xcolor}

\newcommand{\eat}[1]{}

\title{Forecasting in Multivariate Irregularly Sampled Time Series with Missing Values}

\author{
  Shivam Srivastava\thanks{Work done at IBM Almaden Research Center} \\
  Department of Computer Science\\
  University of Massachusetts Amherst\\
  Amherst, MA 01002 \\
  \texttt{shivam@cs.umass.edu} \\
   \And
 Prithviraj Sen \\
  IBM Almaden Research Center\\
  San Jose, CA \\
  \texttt{senp@us.ibm.com} \\
  \And
 Berthold Reinwald \\
  IBM Almaden Research Center\\
  San Jose, CA \\
  \texttt{reinwald@us.ibm.com} \\
}

\begin{document}
\maketitle

\begin{abstract}
Sparse and irregularly sampled multivariate time series are common in clinical, climate, financial and many other domains. Most recent approaches focus on classification, regression or forecasting tasks on such data. In forecasting, it is necessary to not only forecast the right value but also to forecast when that value will occur in the irregular time series. In this work, we present an approach to forecast not only the values but also the time at which they are expected to occur.
\end{abstract}


\section{Introduction}
\eat{Missing values and irregularly collected samples commonly occur in multivariate time series data, making forecasting tasks difficult. Much of the methodology for time-series forecasting assumes that signals are measured systematically at fixed time intervals. However, much real world data can be sporadic, i.e., the signals are sampled irregularly and not all signals are measured each time. A typical example is sensors reporting climate variables like temperature, where not all sensors report at the same time and moreover there is no regularity in each sensor's reports. Modeling this becomes challenging as standard models like recurrent neural networks assume regularly sampled data.\\}

A number of time series applications naturally produce missing values. Examples include electronic health records (EHR) consisting of patient visits where every possible test is not reported during every visit perhaps due to the costs of running healthcare tests. Other examples include climate/weather data, ecology, and astronomy. Historically, such missing data has been handled by imputation which requires choosing a function to impute with, followed by modeling the imputed time series with well understood time series modeling approaches \cite{boxjenkins:book08}. Given that modern data collection techniques allow for collection of voluminous data with relative ease, approaches involving imputation have fallen out of favor especially for cases that would result in an explosion in size due to imputing the missing values. Due to such reasons, a number of recent works have explored how to model time series data in its raw form, without imputation.

\begin{figure}
    \centering
    \includegraphics[width=0.6\linewidth]{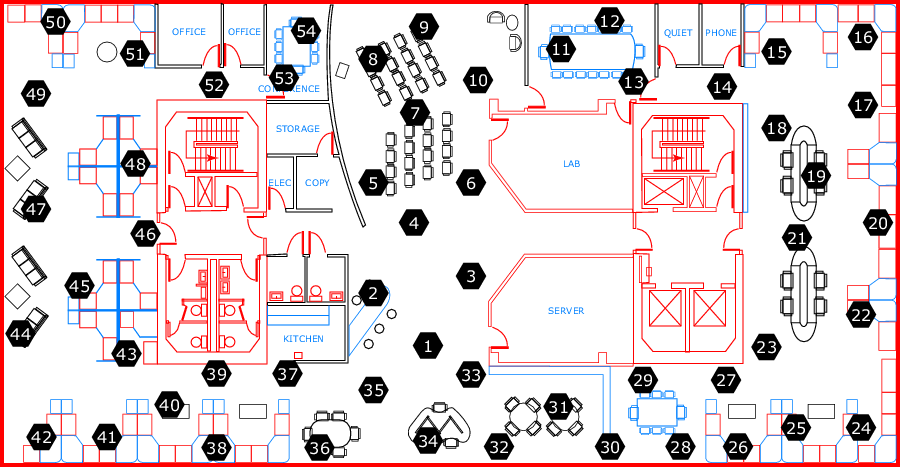}
    \caption{Sensor layout across the lab.}
    \label{labdata}
\end{figure}

Due to the reasons described above, recent works have explored a variety of approaches ranging from recurrent neural networks to attention mechanisms to learn from time series data with missing values. Given that patient EHR has been utilized as the primary benchmark in most of the recent works, certain commonalities exist. For instance, while EHR may contain multiple healthcare indicators in each visit it is unlikely that one patient's healthcare indicators will correlate with another patient's healthcare indicators. Thus, in almost all of the recent works, it is easy to determine which variables are reported at the same (irregularly sampled) timestep and which are not. Another commonality is to test the approaches on downstream classification tasks (e.g., does the patient suffer from diabetes?) rather than standard time series related tasks such as forecasting. We tackle the much more challenging setting where different spatially located sensors record multiple physical attributes whose values may be correlated with each other depending on how far they are located. Figure \ref{labdata} shows how the sensors are located across the lab. Each sensor reports four attributes: temperature, humidity, light and voltage of the battery in the sensor. Each time a sensor reports, it reports the value of all four attributes, but different sensors report its attributes independently. In other words, it is not immediately clear which sensors' reporting patterns are correlated. Moreover, sensors can break. For instance, towards the end of the reporting period the data (available at \url{http://db.csail.mit.edu/labdata/labdata.html}) contains unrealistically high temperature values exceeding 100$^\circ$C.

In this paper, we show how to model the above setting using recurrent neural networks. As mentioned above, our setting is more challenging than the standard setting of time series with missing data on at least two counts: 1) since we are not given which multi-variate attributes are correlated with each other and need to learn such correlations, and 2) because in our setting sensors can break and may report erroneous values. Moreover, we evaluate the proposed methods on downstream forecasting tasks as opposed to classification tasks that may have been the focus of previous works. In terms of results, we show that a fairly simple model outperforms GRU-D [Che et al., 2018], a state-of-the-art technique for modeling time series data with missing values (see \cite{RiseDise} for a comparison of various techniques where GRU-D turned out to be one of the strongest baselines).

\eat{Significant effort has been expended in applying existing techniques and developing new ones to handle data with these characteristics, most significantly in the areas of Gaussian Processes (Schulam and Arora, 2016) and deep learning (Lim and van der Schaar, 2018). These techniques are typically tailored to address classification, regression or forecasting. In forecasting, these techniques do not address what future time that forecast is made for. We present a general approach where the forecast not only has the values but also the time at which the values are expected in future.}

\section{Related Work}
\label{sec:headings}
Machine learning has a long history in time series modelling (Mitchell, 1999; Gers et al., 2000; Wang et al., 2006; Chung et al., 2014). However, the recent massive real-world data collections increase the need for models capable of handling such complex data. Often this data is multivariate and is collected at irregular intervals leading to missing values. The above mentioned techniques do not handle multivariate irregularly sampled data with missing values. Handling this sporadic data is the main difficulty. \\

To address irregular sampling, a popular approach is to recast observations into fixed duration time bins. However, this representation results in missing observations both in time and across features dimensions. This makes the direct usage of neural network architectures tricky. To overcome this issue, the main approach consists in some form of data imputation and jointly feeding the observation mask and times of observations to the recurrent network (Che et al., 2018; Choi et al., 2016a; Lipton et al., 2016; Du et al., 2016; Choi et al., 2016b; Cao et al., 2018). This approach strongly relies on the assumption that the network will learn to process true and imputed samples differently. Despite some promising experimental results, there is no guarantee that it will do so. Some researchers have tried to alleviate this limitation by introducing a more meaningful data representation for sporadic time series (Rajkomar et al., 2018; Razavian \& Sontag, 2015; Ghassemi et al., 2015). \cite{RiseDise} evaluates a number of approaches described above by casting them as specific instances of the RISE framework (\underline{R}ecursive \underline{I}nput and \underline{S}tate \underline{E}stimation) which allows an internal hidden state vector $h_i$ to be computed (usually by some variant of the RNN family), a transformed hidden state vector $\hat{h}_i$ (usually an older hidden state is decayed depending on how long ago it was computed), and a transformed input $\hat{x}_i$ that allows for the inclusion of some imputation mechanism. One of its primary observations is that the current techniques lack mechanisms to handle long-horizon forecasts. Their experiments are conducted on univariate time series (patients with glucose measurements made every 5 minutes) and 2-variable tasks (PM$_{2.5}$ and PM$_{10}$ measured every hour across monitoring stations in Beijing).\\

Among slightly older works, RETAIN \cite{choi:nips16} combines attention with RNNs to address missing data in EHR while retaining some semblance of interpretability so doctors can gain some trust into the underlying modeling mechanism and predictions made. More notably, they run their RNNs in reverse, i.e., start at the most recent observed values iterating to the past, and also they do not make use of the time lags among the observations. Doctor AI \cite{choi:mlhc16} uses various codes available in the medical domain (e.g., diagnostic codes such as ICD) to feed into an RNN to construct a hidden representation of the patient's status. They describe an approach to learn embeddings for the input codes using word2vec \cite{mikolov:nips13} to feed to the RNN, besides feeding in the delta time lag since the last observation. T-LSTM \cite{baytas:kdd17}, like GRU-D, also proposes to discount the state of the RNN depending on how long ago the previous observation was made by dividing the RNN hidden state vector into a long-term component and a short-term component.

\par

Others have addressed the missing data problem with generative probabilistic models. Among those, (multitask) Gaussian processes (GP) are the most popular by far (Bonilla et al., 2008). They have been used for smart imputation before a RNN or CNN architecture (Futoma et al., 2017; Moor et al., 2019), for modelling a hidden process in joint models (Soleimani et al., 2018), or to derive informative representations of time series (Ghassemi et al., 2015). GPs have also been used for direct forecasting (Cheng et al., 2017). However, they usually suffer from high uncertainty outside the observation support, are computationally intensive and learning the optimal kernel is tricky. Neural Processes, a neural version of GPs, have also been introduced by Garnelo et al. (2018).\\

Most recently, the seminal work of Chen et al. (2018) suggested a continuous version of neural networks that overcomes the limits imposed by discrete-time recurrent neural networks. Coupled with a variational auto-encoder architecture (Kingma \& Welling, 2013), it proposed a natural way of generating irregularly sampled data. However, it transferred the difficult task of processing sporadic data to the encoder, which is a discrete-time RNN. Related auto-encoder approaches with sequential latents operating in discrete time have also been proposed (Krishnan et al., 2015, 2017). These models rely on classical RNN architectures in their inference networks, hence not addressing the sporadic nature of the data. What is more, if they have been shown useful for smoothing and counterfactual inference, their formulation is less suited for forecasting. Importantly, the ability of the GRU to learn long-term dependencies is a significant advantage.\\

\textit{Most of these models focus on classification whereas we focus on forecasting the values along with the time of their forecast.}

\section{Methods}

In this section we first formally define the problem statement, then discuss some preliminaries and finally the proposed model.

\subsection{Problem Statement}
\eat{We consider the general problem of one step forecasting on $N$ sporadically observed $D$-dimensional time series.} We consider the general problem of one step forecasting for data from $D$ sensors where $L$ climate longitudinal variables can potentially be measured. We pick a single longitudinal variable ($L$=1). We use an autoregression hyperparameter $AR$. $N$ denotes the length of the sequence data (number of time steps), and the data at each $i \in \{1,...,N\}$ constitutes a time series. The $N$ time series are obtained by taking $AR$ steps of the $D$-dimensional data where each consecutive sequence overlaps on $AR-1$ common steps. Each time series $i$ is measured at $AR$ time points specified by a vector of observation times $t_i \in \mathbb{R_{+}}^{AR}$. Let $s_{t}$ $\in \mathbb{R_{+}}$ denote the time-stamp when the $t^{th}$ observation is obtained and we assume that the first observation is made at time-stamp 0 (i.e., $s_1$=0). The values of these observations are specified by a matrix of observations $x_i \in \mathbb{R}^{AR \times D}$, an observation mask $m_i \in \{0,1\}^{AR \times D}$ and a matrix $\Delta_i \in \mathbb{R_{+}}^{AR \times D}$ specifying the time difference between each observation of a variable. To be more specific, for a given $D$-dimensional time series $x$ of length $AR$, we have:
\[
    m_{t}^{d}=\left\{
                \begin{array}{ll}
                  1, \hspace{0.5cm} \text{if $x_{t}^{d}$ is observed}\\
                  0, \hspace{0.5cm} \text{otherwise}
                \end{array}
              \right.
\]

\[
    \Delta_{t}^{d}=\left\{
                \begin{array}{ll}
                  s_{t} - s_{t-1} + \Delta_{t-1}^{d}, \hspace{0.2cm} t $>$ 1, m_{t-1}^{d} = 0\\
                  s_{t} - s_{t-1}, \hspace{1.4cm} t $>$ 1, m_{t-1}^{d} = 1\\
                  0, \hspace{2.6cm} t = 1
                \end{array}
              \right.
\]
where both $m_{t}^{d}$ and $\Delta_{t}^{d}$ are scalars and stand for the respective values of the $d^{th}$ dimension at time step $t$.
In this paper, we are interested in time series forecasting, where we predict the value $v_n \in \mathbb{R}^{D}$ given the time series data $\mathcal{D}$, where $\mathcal{D} = \{(x_n, \Delta_n, m_n)\}_{n=1}^{N}$.

\subsection{Preliminaries: The Gated Recurrent Unit}

Recurrent Neural Networks (RNNs), such as Long Short-Term Memory (LSTM) \cite{hochreiter1997long} and Gated Recurrent Unit (GRU) \cite{cho2014learning}, have many useful properties such as strong prediction performance as well as the ability to capture long-term temporal dependencies (like seasonality) and variable-length observations. They have shown to achieve the state-of-the-art results in many applications with time series or sequential data. RNNs for missing data have been studied in the past \cite{DBLP:journals/corr/ChePCSL16}, \cite{bengio1996recurrent}, \cite{kim2018temporal}, \cite{parveen2002speech}. \\
\noindent
Given our setting, these RNNs are ideal for time series forecasting. The sensor data has seasonality and trend in addition to irregular samples with missing values, so we need a model which can exploit the correlation between the variables and also capture the seasonality. The recursive nature of RNNs allow it to do both. As the parameters are shared across the time steps, the number of parameters is relatively small. We use the Gated Recurrent Units (GRUs) for our experiments but similar techniques are also valid for other variants of RNNs.

\begin{figure}[H]
  \centering
  \includegraphics[width=0.3\linewidth]{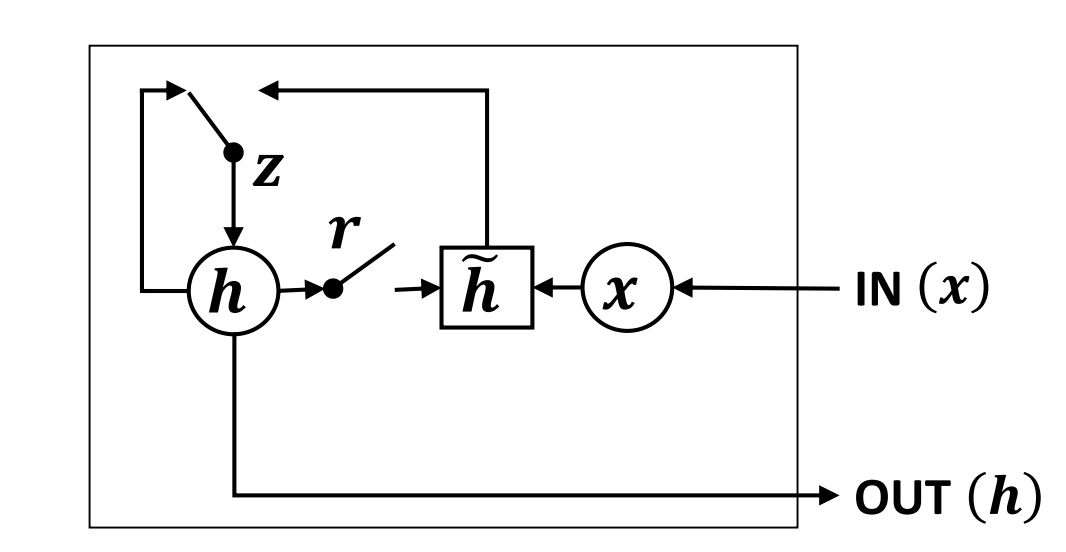}
  \caption{GRU}
  \label{fig:GRU}
\end{figure}
Figure \ref{fig:GRU} shows the structure of GRU. For each j-th hidden unit, GRU has a reset gate $r_t^{j}$ and an update gate $z_t^{j}$ to control the hidden state $h_t^{j}$ at each time $t$. The update functions are:\\

\[ r_t = \sigma(W_{r} x_{t} + U_{r} h_{t-1} + b_{r})\]
\[ z_t = \sigma(W_{z} x_{t} + U_{z} h_{t-1} + b_{z})\]
\[ \tilde{h_{t}} = tanh(Wx_t + U(r_t * h_{t-1}) + b)\]
\[ h_{t} = (1-z_{t}) * h_{t-1} + z_{t}*\tilde{h_t}\]

where matrices $W_z$, $W_r$, $W$, $U_z$, $U_r$, $U$ and vectors $b_z$, $b_r$, $b$ are model parameters. Here $\sigma$ represents element-wise sigmoid function, and $*$ stands for element-wise multiplication. It is assumed in this setup that all the inputs are observed and hence the missing values need to be either explicity or implicity imputed. Depending on the task, either classification or regression, an appropriate last layer, sigmoid/soft-max or dense respectively, is applied on the output of GRU at the last time sep.
\eat{\subsection{Data Preparation}}

\subsection{Proposed Model}

Inspired from GRU-Simple \cite{che2018recurrent}, our model takes $AR$ consecutive time-steps from the $t_i$ vector with the corresponding $m_i$, $x_{i}$ and $\Delta_{i}$ concatenated as the input to the GRU. The target vector is the concatenation of $x_i$ and $\Delta_i$ for the next time-step. More specifically, 
\[ x_i = [x_i; m_i; \Delta_i] \]

The missing values are imputed using forward imputation. More formally, for a given sequence x,
\[x_t^{d} = m_{t}^{d}x_t^{d} + (1 - m_{t}^{d})x_{t^{'}}^{d}\]
where $t^{'} < t$ is the last time the $d$th variable was observed.

\begin{figure}[H]
  \centering
  \includegraphics[width=0.55\linewidth]{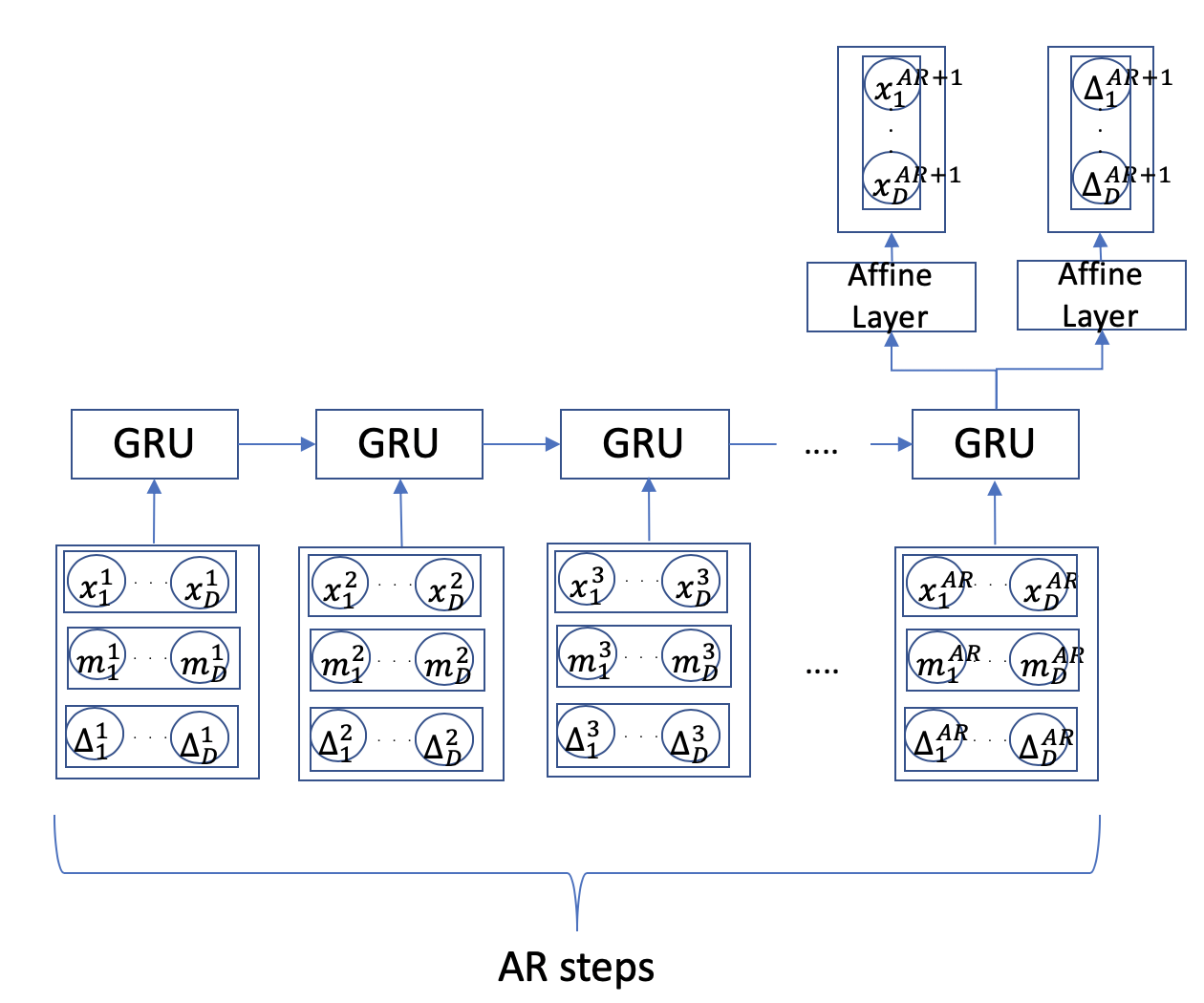}
  \caption{Proposed Forecasting Model ($L$=1)}
  \label{fig:fig1}
\end{figure}

We differ from GRU-Simple \cite{che2018recurrent} in three aspects:
\begin{itemize}
    \item We focus on forecasting instead of classification.
    \item We not only forecast the values but also the time at which they are expected to occur.
    \item We use a custom loss function which optimizes only over the present values.
\end{itemize}
We can consider our neural network as a function $f_{\theta}$ parameterized on $\theta$. Then the output $\hat{y}_{n}$ = $f_{\theta}(x_n)$. The loss for the $i^{th}$ input sequence is \[l_{i} = m_{i} * L(y_{i}, \hat{y}_{i})\] where $L$ is chosen to be HuberLoss in order to minimize the Mean Absolute Error (MAE). The same loss function is used for the delta prediction layer too. Figure \ref{fig:fig1} shows the model.
\label{sec:others}

\section{Evaluation}
This section provides details about the dataset and the results obtained using different techniques. Later, this section also describes the properties and shortcomings of the proposed approach.
\subsection{Dataset}
We use Intel's Berkeley Lab data from 2005 which consists of 58 sensors measuring 4 physical attributes: temperature, humidity, voltage and light. Of these 4, we select temperature for a subset of 4 sensors (sensor ID: 1, 2, 3 and 4) for simplicity and faster experiments. The original data at microsecond granularity is compressed to a second level granularity. The details for the temperature variable in the dataset are given below.
\begin{table}[H]
\centering
\begin{tabular}{|c|c|c|l|l|}
\hline
\textbf{Timesteps} & \textbf{\#Sensors} & \textbf{Temperature Avg ($^{\circ}$C)} & \textbf{Delta Avg (s)} & \textbf{Sparsity} \\ \hline
968,535            & 58                 & 39                                 & 62                     & 0.04117           \\ \hline
\end{tabular}
\end{table}

\begin{figure}[H]
  \centering
  \includegraphics[width=0.8\linewidth]{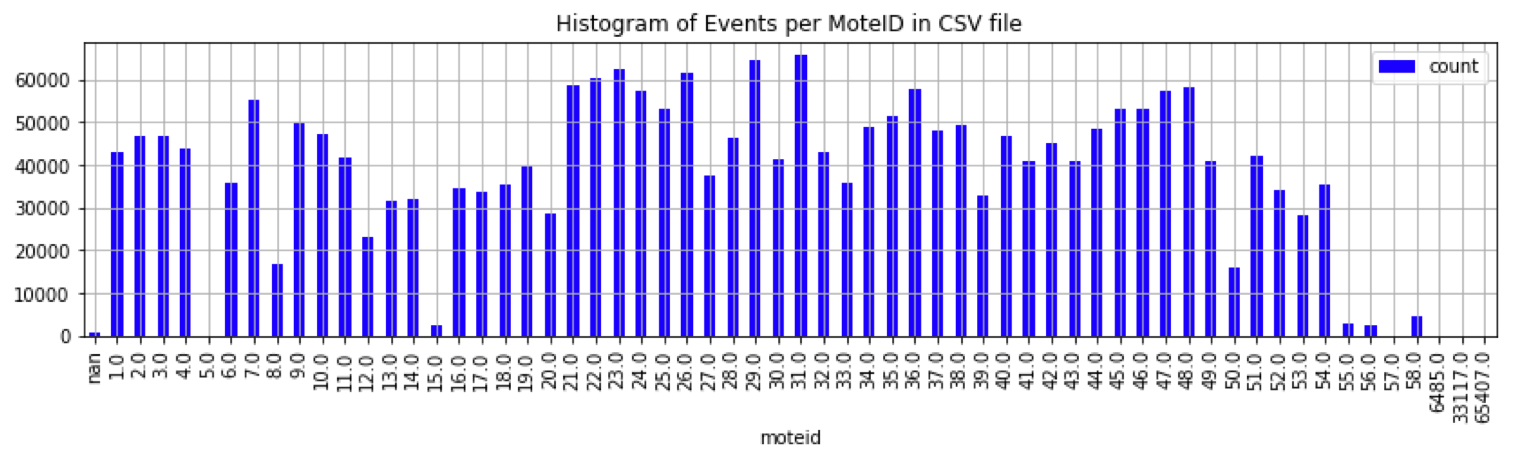}
  \caption{Event frequency across all sensors in the dataset}
  \label{fig:fig2}
\end{figure}

Details of the subset used for experiments:
\begin{table}[H]
\centering
\begin{tabular}{|c|c|c|l|l|}
\hline
\textbf{Timesteps} & \textbf{\#Sensors} & \textbf{Temperature Avg ($^{\circ}$C)} & \textbf{Delta Avg (s)} & \textbf{Sparsity} \\ \hline
167,875            & 4                  & 38.77                              & 61.64                  & 0.3587            \\ \hline
\end{tabular}
\end{table}

\begin{figure}[h]       
\centering
    \mbox{\includegraphics[width=60mm]{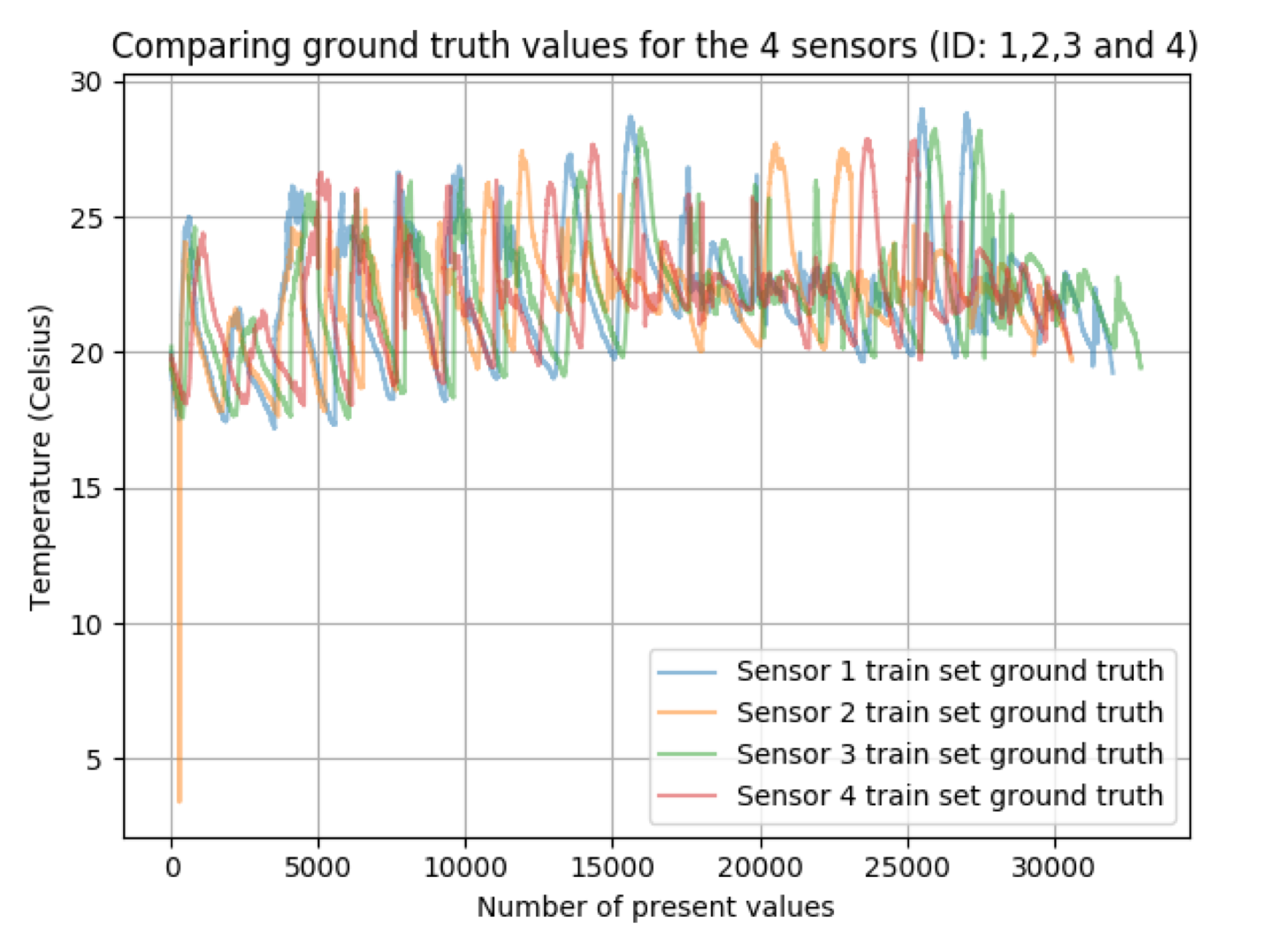}}   
    \hspace{30px}
    \mbox{\includegraphics[width=60mm]{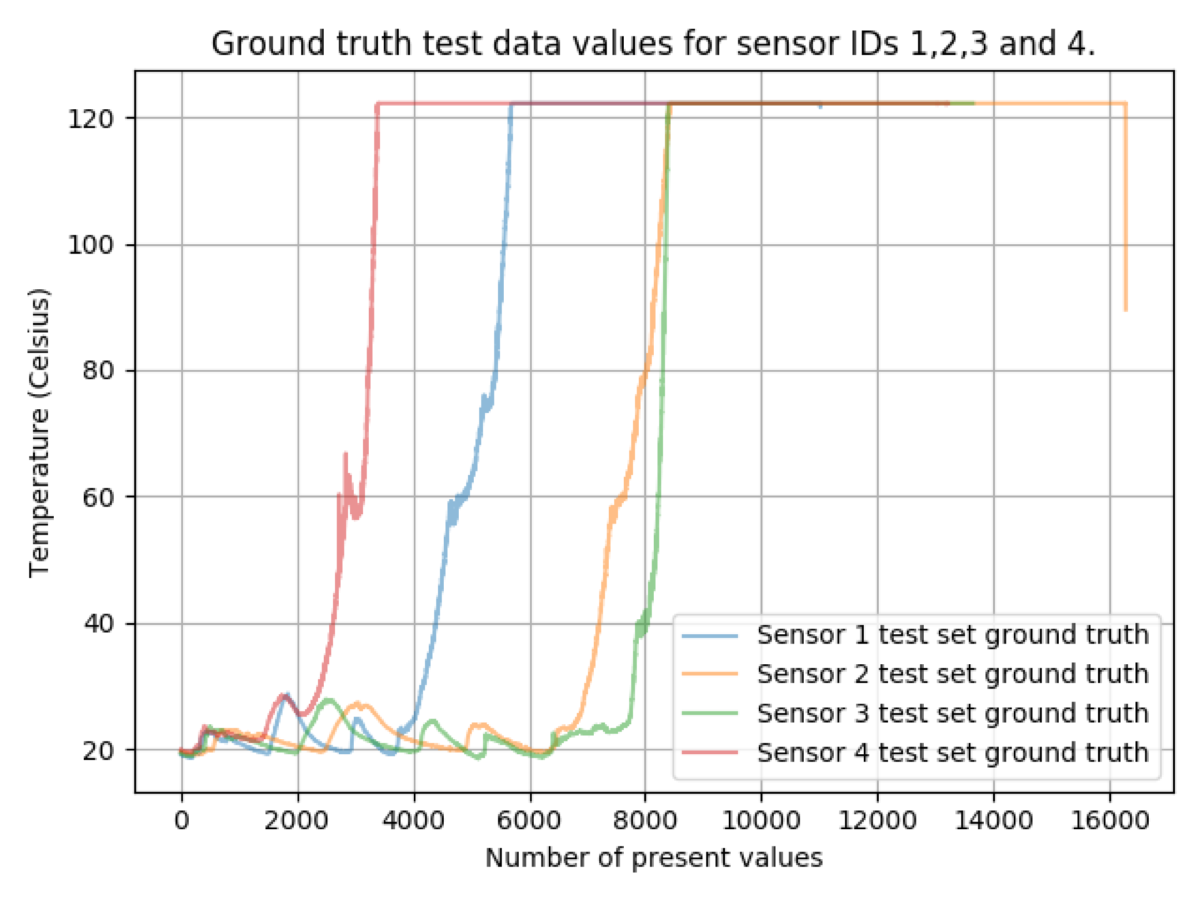}}
    \caption{Training Set [left] and Test Set (last 30\% of the data)  [right] }
    \label{fig:fig4}
\end{figure}
\FloatBarrier


Test set is created by holding out the last 30\% of the dataset. See Figure \ref{fig:fig4} for the plot.

\subsection{Results}
\begin{figure}[H]
  \centering
  \includegraphics[width=0.45\linewidth]{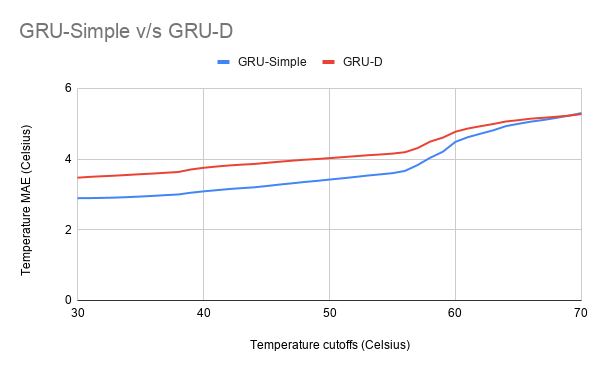}
  \caption{GRU-Simple v/s GRU-D comparison on temperature cut-off in the test set.}
  \label{fig:fig5}
\end{figure}

We compare our technique with GRU-D and GRU-ODE-Bayes on the above dataset.
\begin{table}[H]
\centering
\begin{tabular}{|c|c|c|}
\hline
\textbf{Technique} & \textbf{Temperature MAE ($^{\circ}$C)} & \textbf{Delta MAE (s)} \\ \hline
GRU-Simple         & 2.89                         & 18.2                   \\ \hline
GRU-D              & 3.47                         & -                      \\ \hline
GRU-ODE-Bayes      & 3.58                         & -                      \\ \hline
\end{tabular}

\caption{Error values on test set with a maximum temperature cut-off at 30$^{\circ}$C}
\end{table}
Figure \ref{fig:fig5} compares GRU-D and GRU-Simple for different temperature cut-offs on the test set showing that GRU-Simple performs better on the part of the test set without anomalies.

\FloatBarrier



\begin{figure}[h]       
\centering
    \mbox{\includegraphics[width=53mm]{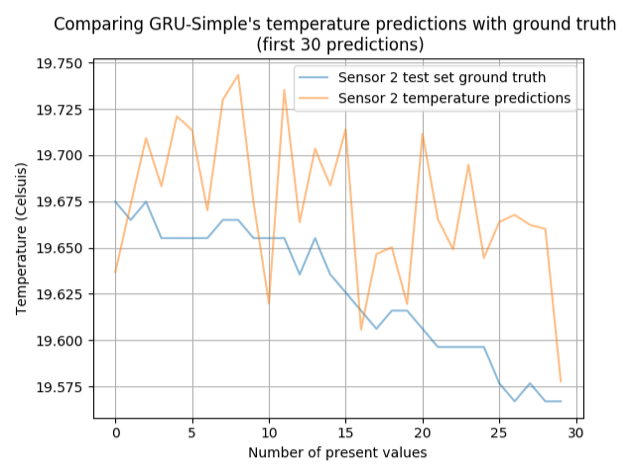}}   
    \mbox{\includegraphics[width=53mm]{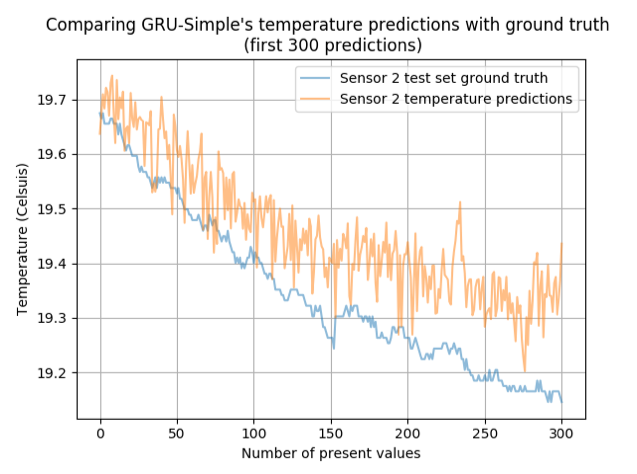}}
    \mbox{\includegraphics[width=53mm]{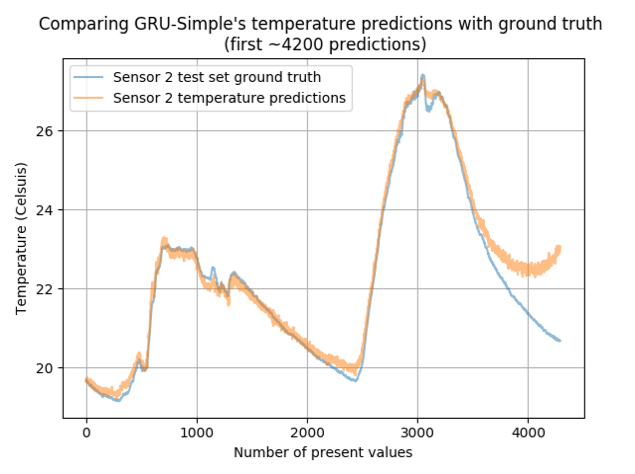}}
    \caption{GRU-Simple Temperature predictions at different detail levels}
    \label{GRUSimple-Temperature}
\end{figure}
\FloatBarrier

\begin{figure}[h]       
\centering
    \mbox{\includegraphics[width=53mm]{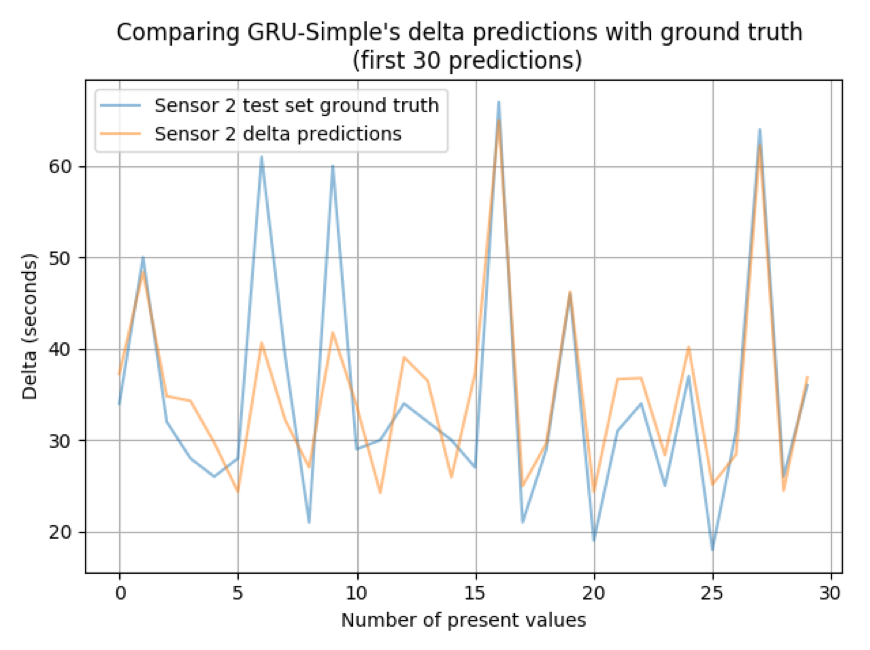}}   
    \mbox{\includegraphics[width=53mm]{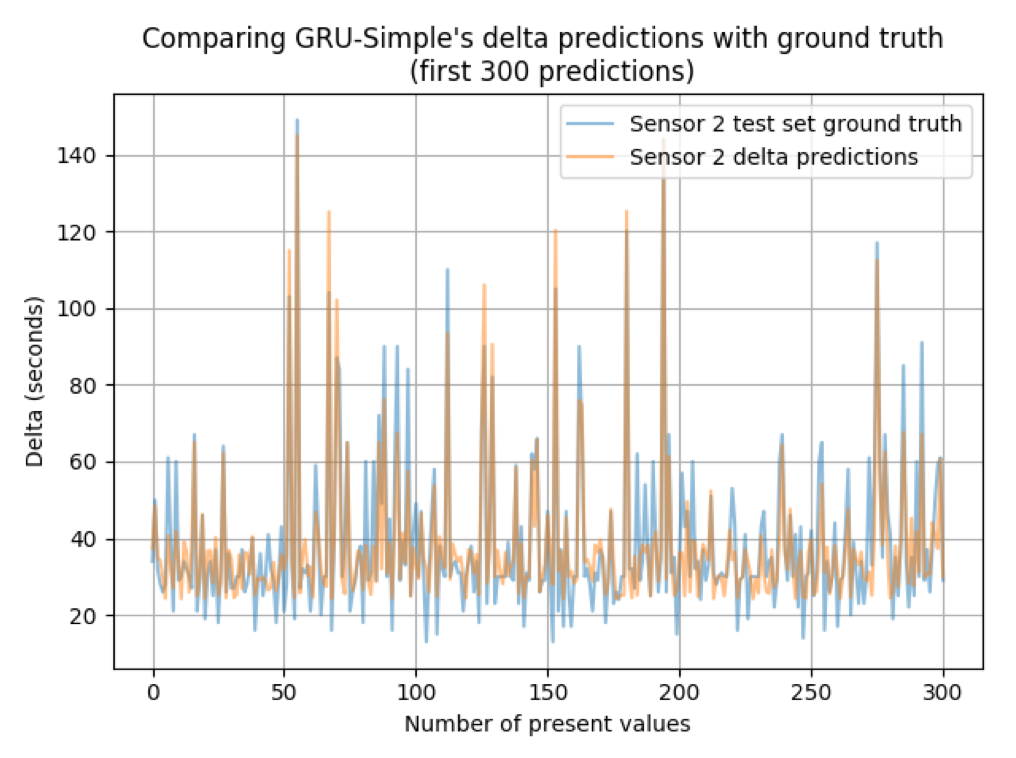}}
    \mbox{\includegraphics[width=53mm]{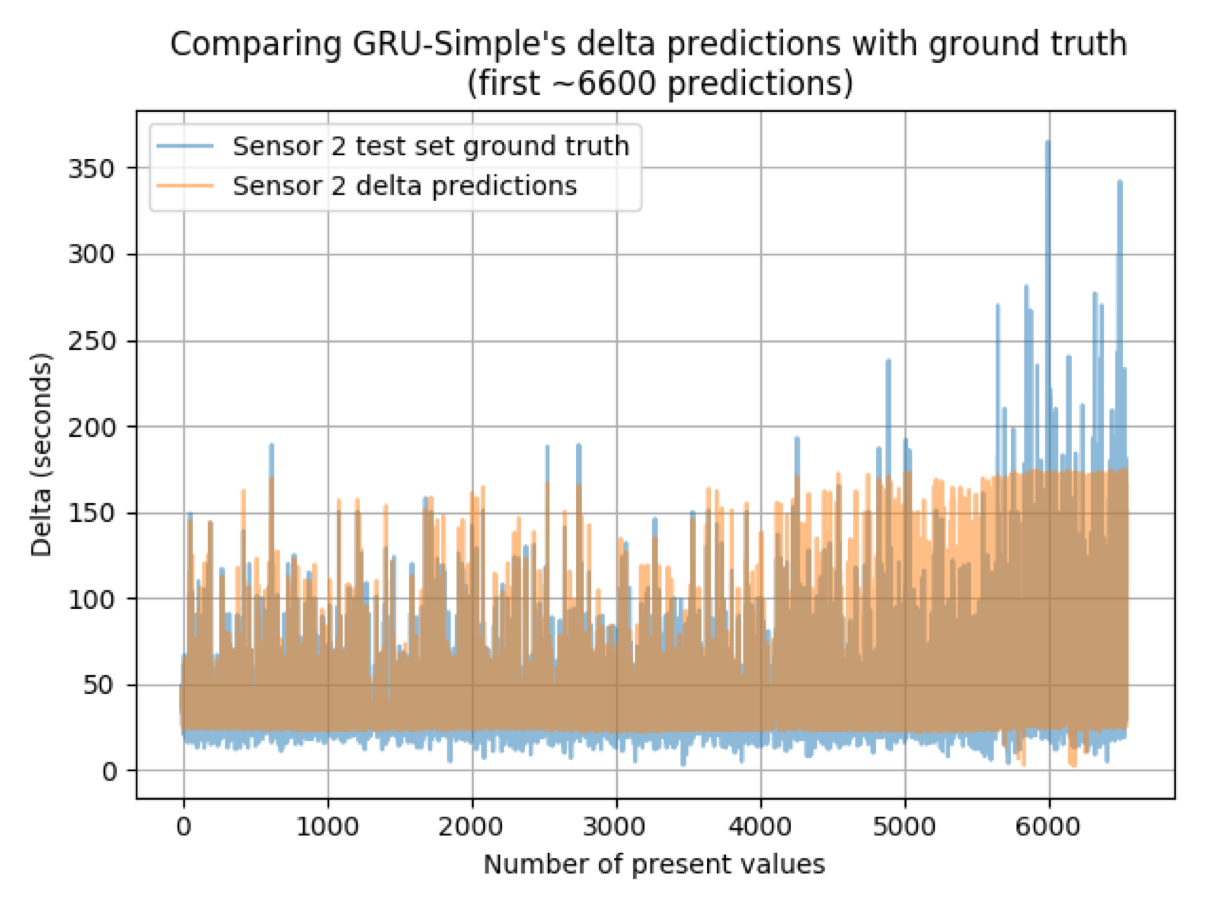}}
    \caption{GRU-Simple Delta predictions at different detail levels}
    \label{GRUSimple-Delta}
\end{figure}
\FloatBarrier

\begin{figure}[h]       
\centering
    \mbox{\includegraphics[width=53mm]{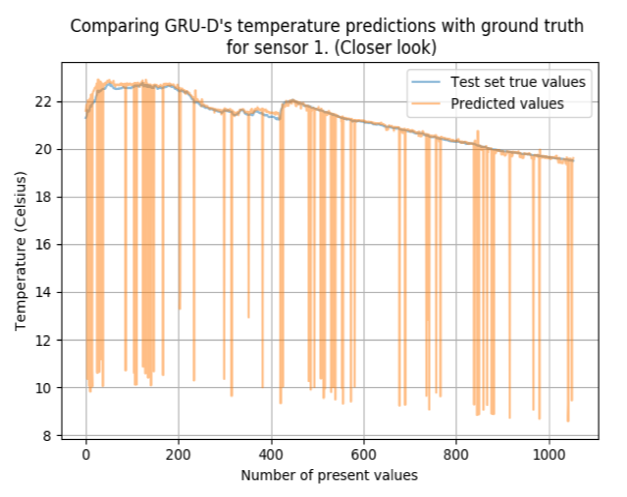}}
    \mbox{\includegraphics[width=53mm]{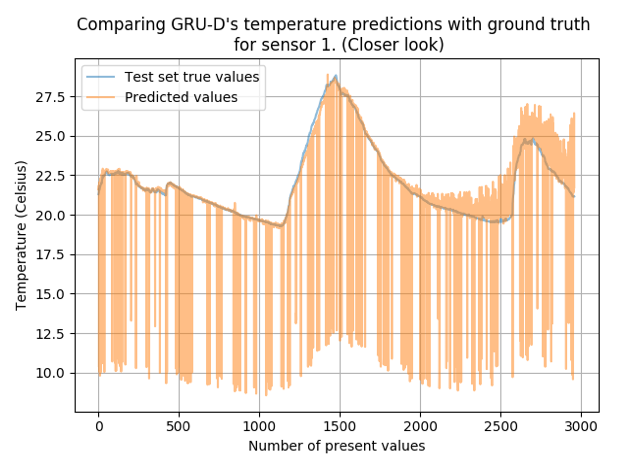}}
    \mbox{\includegraphics[width=53mm]{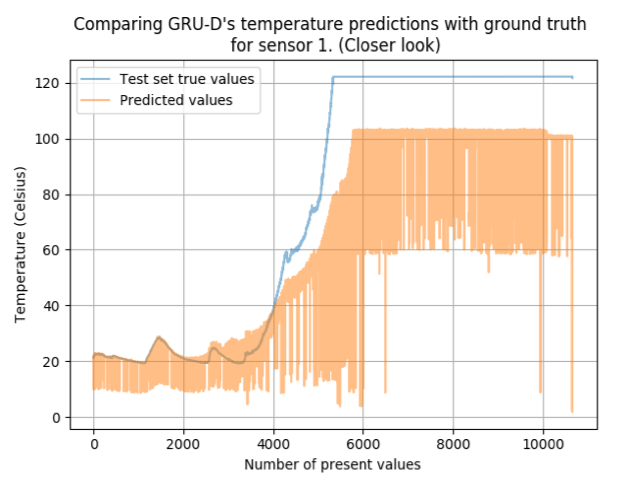}}
    \caption{GRU-D predictions on the test set at different detail levels}
    \label{GRUD-Temperature}
\end{figure}
\FloatBarrier
GRU-D is sensitive to shuffling the dataset. Plots in Figure \ref{GRUD-Temperature} were obtained by training on an unshuffled training set and then evaluating on an unshuffled test set (left shows zoomed in version while right shows zoomed out version). The dataset was also shuffled and tested. The plots obtained were similar. Why certain irregularities occur in these plots is difficult to explain. In the next section let's look at some properties of GRU-Simple.
\subsection{GRU-Simple Properties}
\begin{figure}[h]       
\centering
    \mbox{\includegraphics[width=70mm]{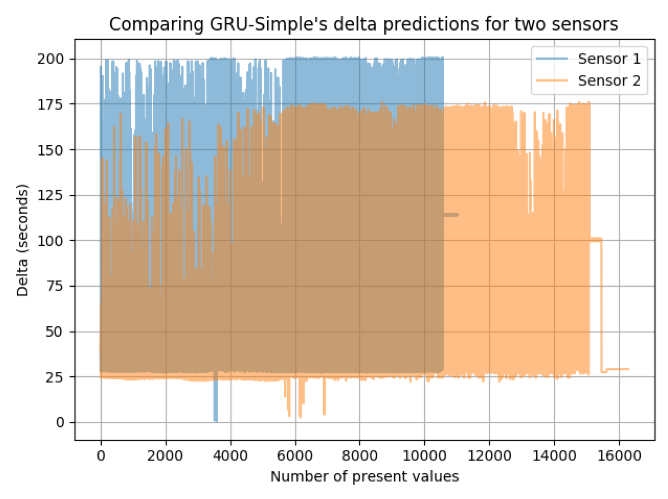}}
    \hspace{30px}
    \mbox{\includegraphics[width=70mm]{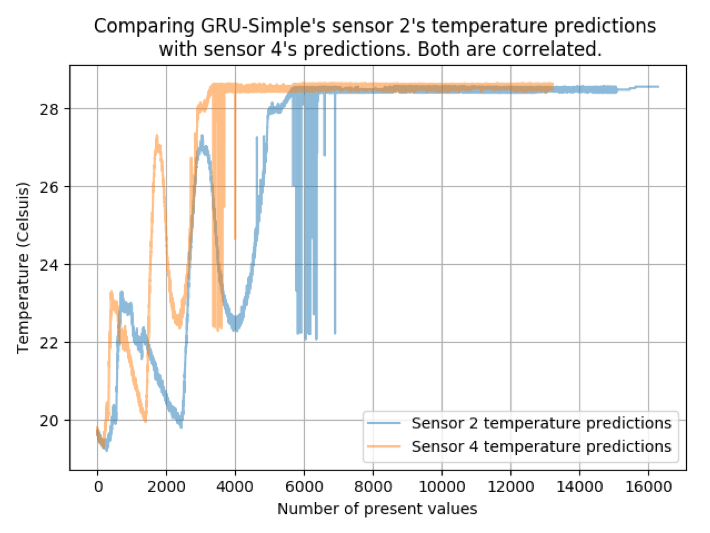}}
    \caption{Correlation between GRU-Simple's predictions for $\Delta$ and temperature.}
    \label{TempDeltaCorrelation}
\end{figure}

Plots in Figure \ref{GRUSimple-Temperature} overlay forecasted temperatures over ground truth temperature while plots in Figure \ref{GRUSimple-Delta} do the same with delta prediction. $\Delta$ values are highly uncorrelated. GRU-Simple performs well despite that. Temperatures for the 4 sensors are highly correlated. The predictions are likewise highly correlated. Figure \ref{TempDeltaCorrelation} shows the plots.
\FloatBarrier

\subsection{Imputation and loss function}
In this section, we justify the imputation and the loss function used. Later we also discuss the problems with GRU-Simple.
\subsubsection{Is imputation important?}
\begin{figure}[h]       
\centering
    \mbox{\includegraphics[width=70mm]{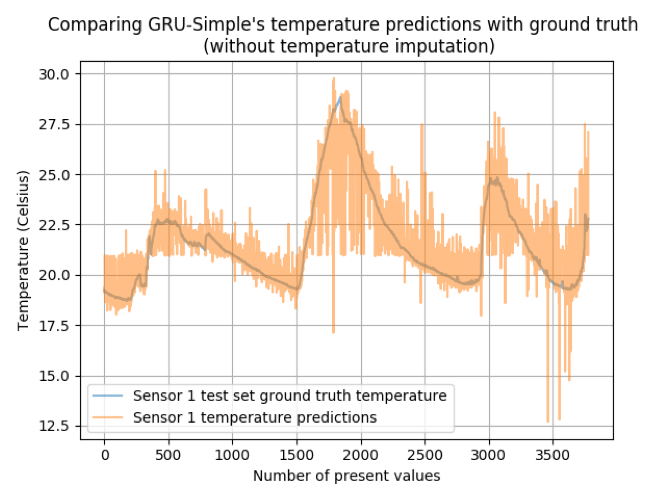}}
    \hspace{30px}
    \mbox{\includegraphics[width=70mm]{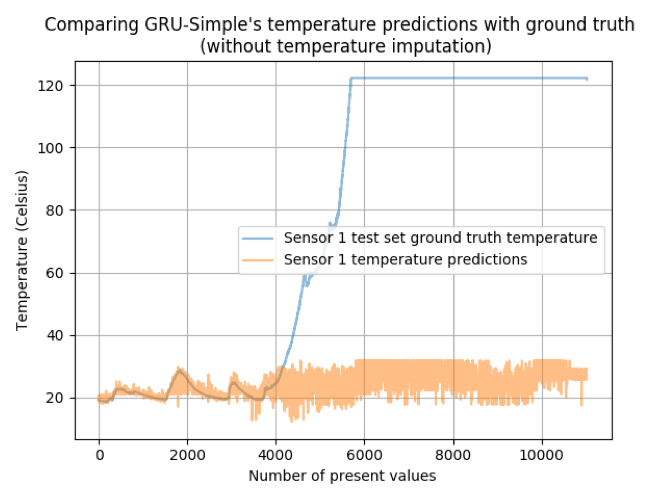}}
    \caption{Importance of imputation}
    \label{Imputation}
\end{figure}

As Figure \ref{Imputation} shows, a reasonable imputation technique is necessary. If not imputed, the missing values are replaced by 0 by default as the data cannot have NaNs as input to the model. In our model, we use forward imputation. The literature for imputation is already replete with various techniques so that is not the focus of our work.
\FloatBarrier

\subsubsection{Is directly optimizing the present values important?}
\begin{figure}[h]       
\centering
    \mbox{\includegraphics[width=70mm]{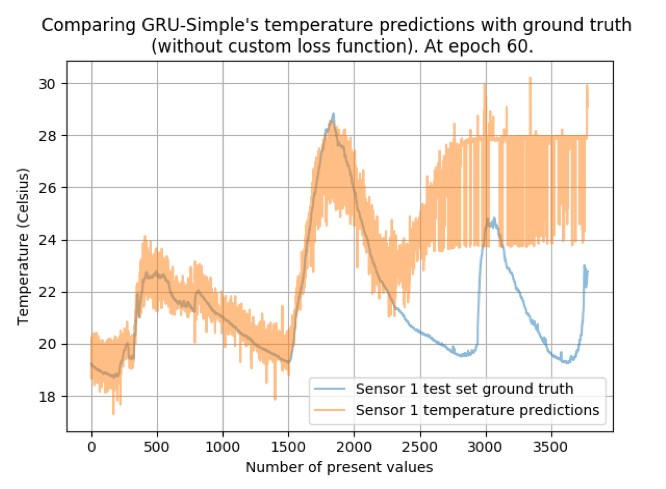}}
    \hspace{30px}
    \mbox{\includegraphics[width=70mm]{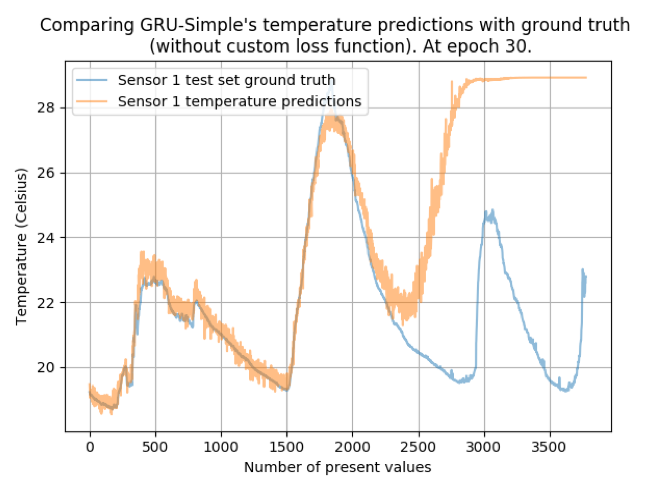}}
    \caption{Importance of the custom loss function}
    \label{CustomLoss}
\end{figure}

As Figure \ref{CustomLoss} shows, it is necessary. The predictions are sometimes jagged, sometimes smooth. Optimizing the present values directly, makes sure the correct values are optimized and the prediction plot gets smoother.
\FloatBarrier

\subsubsection{Problems with GRU-Simple}
\begin{figure}[h]       
\centering
    \mbox{\includegraphics[width=70mm]{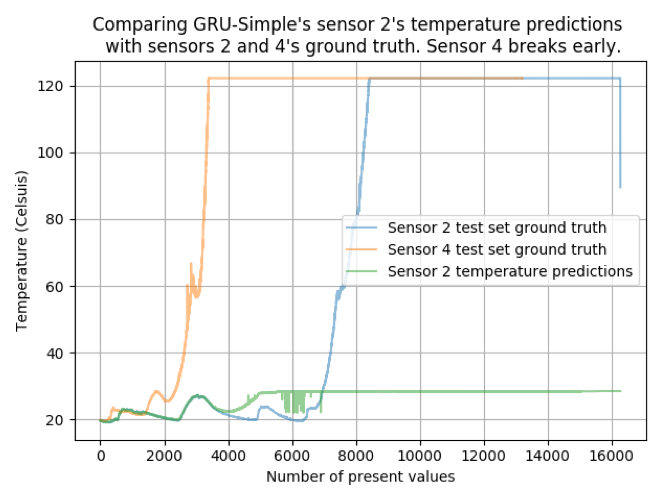}}
    \hspace{30px}
    \mbox{\includegraphics[width=70mm]{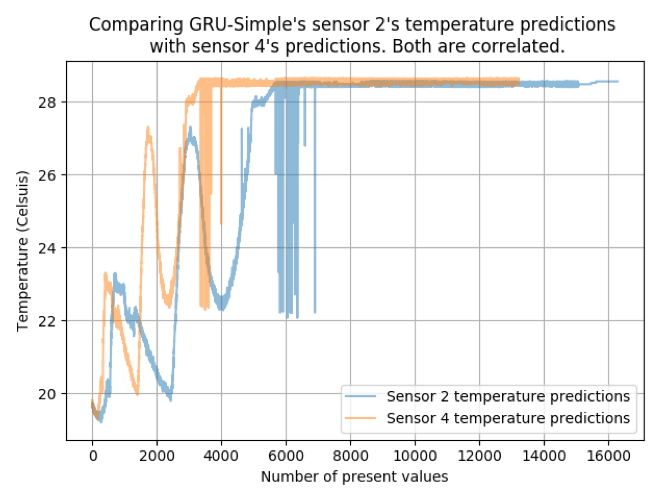}}
    \caption{Problems with GRU-Simple}
    \label{GRU_Simple_Problems}
\end{figure}

GRU-Simple has two problems.
\begin{itemize}
    \item Input of one sensor affects the output of another sensor (Figure \ref{GRU_Simple_Problems} left).
    \item Predictions are highly correlated even if the last AR steps of the two sensors are different (Figure \ref{GRU_Simple_Problems} right).
\end{itemize}

\FloatBarrier
To overcome the above problems we tried a few variations as discussed in the next section.
 
\section{Variations on GRU-Simple}
The motivation behind the variations is to overcome the two shortcomings discussed in the previous section. The next few subsections explain them.

\subsection{BiLayer GRU}
This approach was taken to overcome the high correlation between the outputs. The motivation behind this is to treat temperature as a dependent variable on $\Delta$. The time at which a particular sensor reports a value is random and hence should be treated as an independent variable. Figure \ref{GRU_BiLayer} below shows the model.

\begin{figure}[h]       
\centering
    \mbox{\includegraphics[width=80mm]{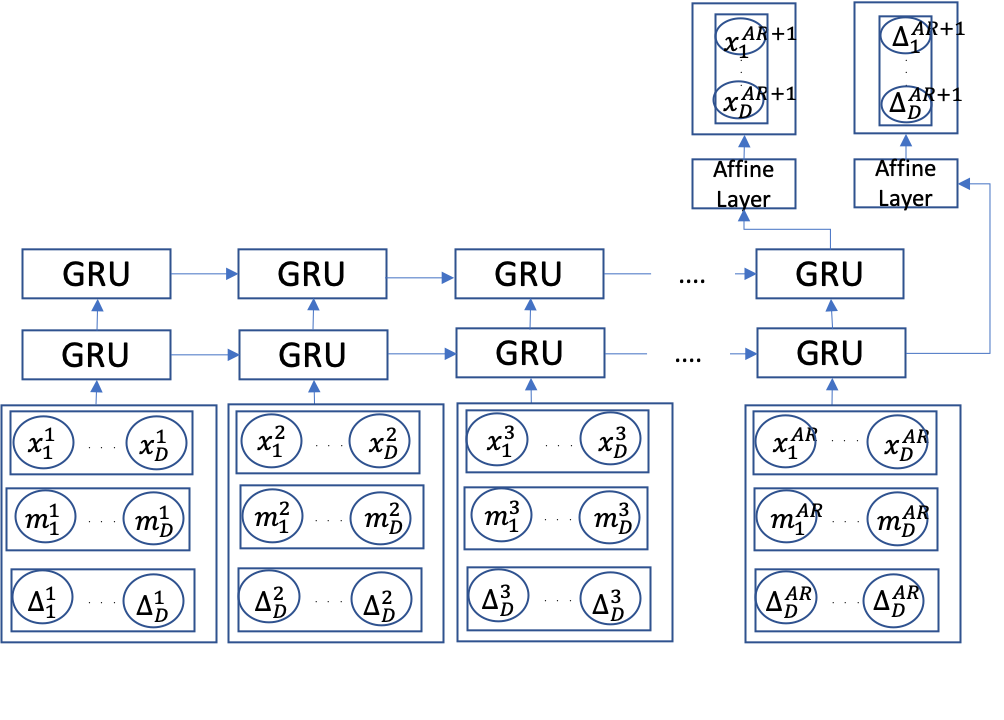}}
    \caption{GRU Bi-Layer}
    \label{GRU_BiLayer}
\end{figure}
\begin{figure}[h]       
\centering
    \mbox{\includegraphics[width=70mm]{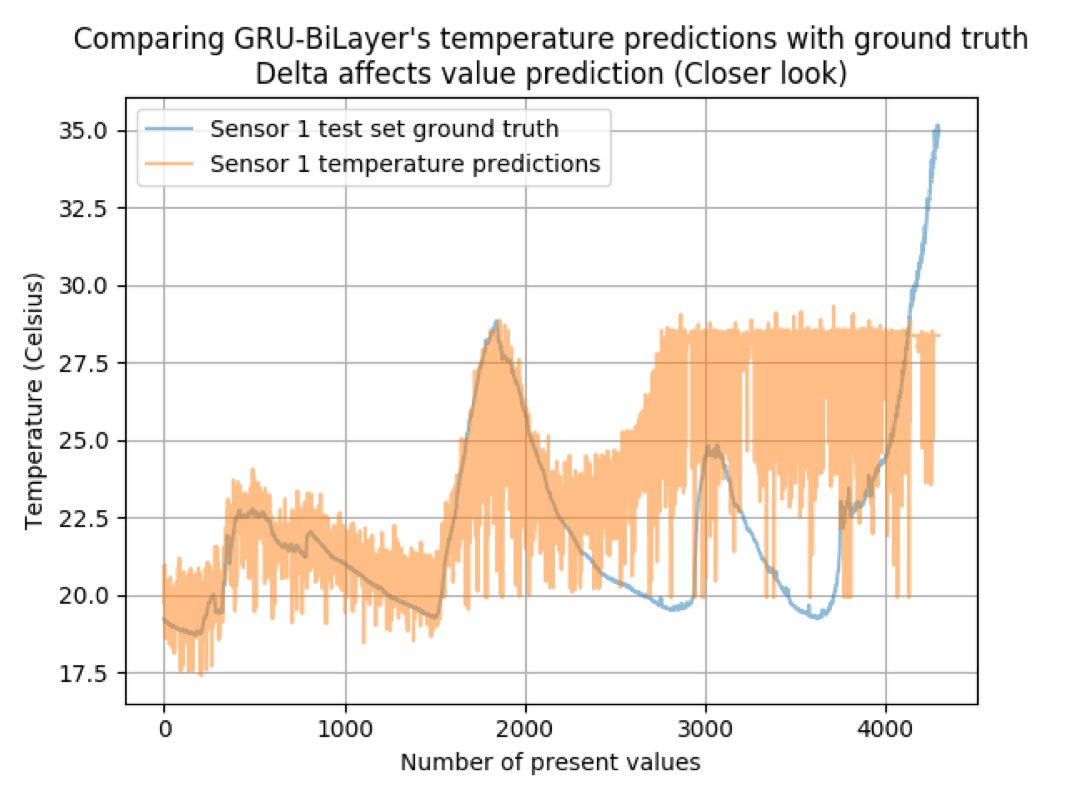}}
    \hspace{30px}
    \mbox{\includegraphics[width=70mm]{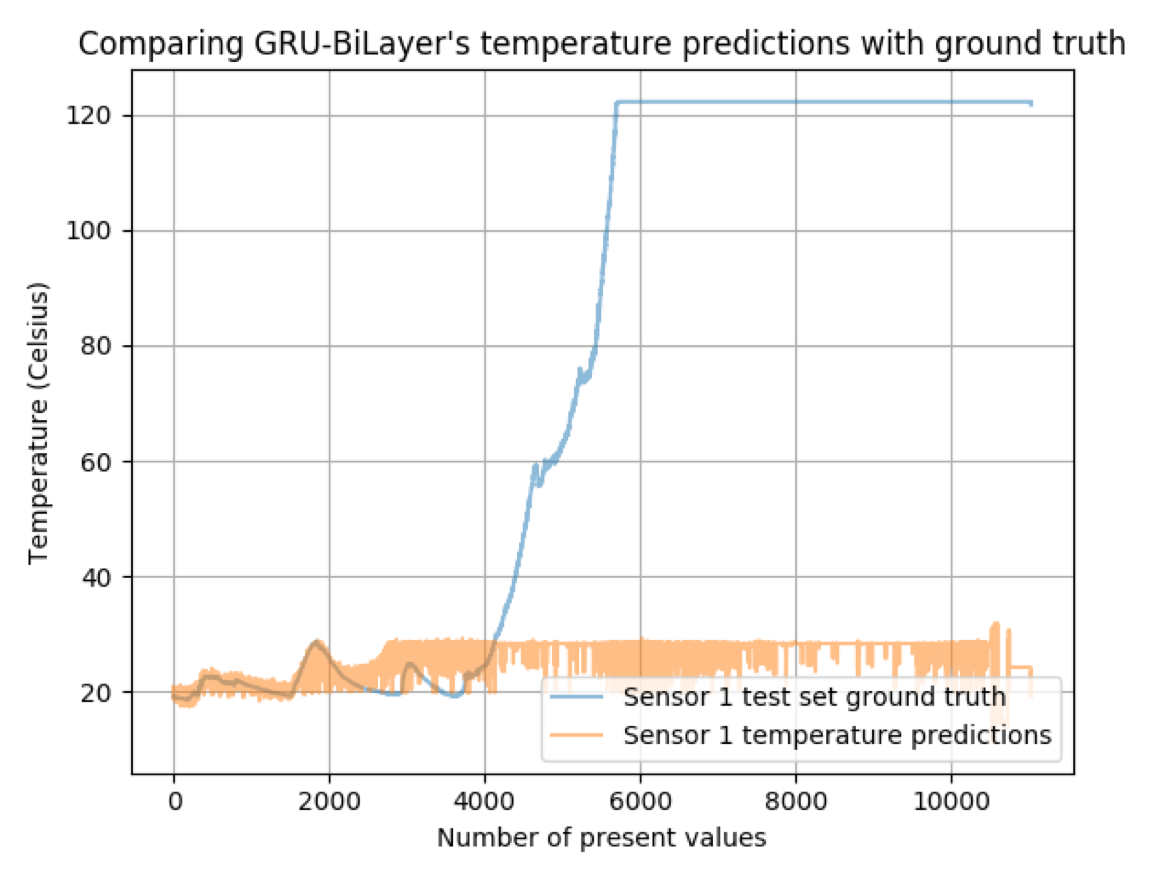}}
    \caption{Problems with GRU-BiLayer}
    \label{GRU_BiLayer_Problems}
\end{figure}

\FloatBarrier
The above approach resulted in temperature predictions which were affected by delta predictions as seen in the next plots (Figure \ref{GRU_BiLayer_Problems}) and hence cannot be used reliably.

\FloatBarrier

\subsection{GRU-Velocity}
The motivation here is to obtain the next temperature prediction using the following equation:
\[x_{AR+1} = x_{AR} + (\frac{dx}{dt})^{AR+1} * \Delta^{AR+1}\]

\begin{figure}[h]       
\centering
    \mbox{\includegraphics[width=90mm]{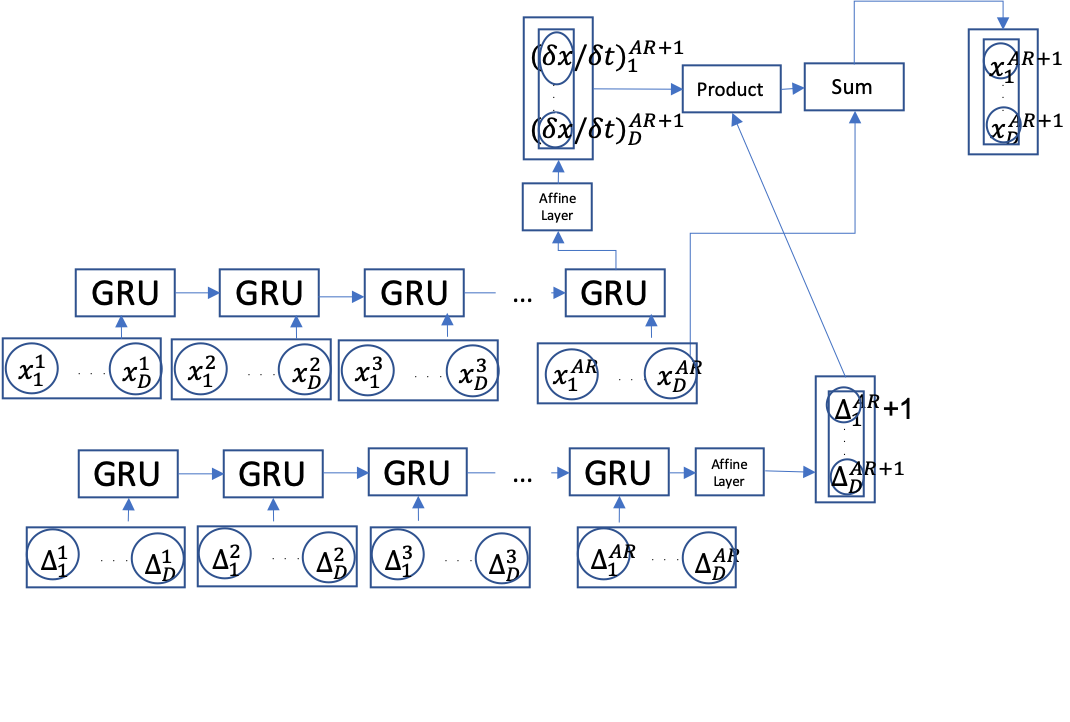}}
    \caption{GRU Velocity}
    \label{GRU_Velocity}
\end{figure}

The architecture is shown in Figure \ref{GRU_Velocity}.

\begin{figure}[h]       
\centering
    \mbox{\includegraphics[width=85mm]{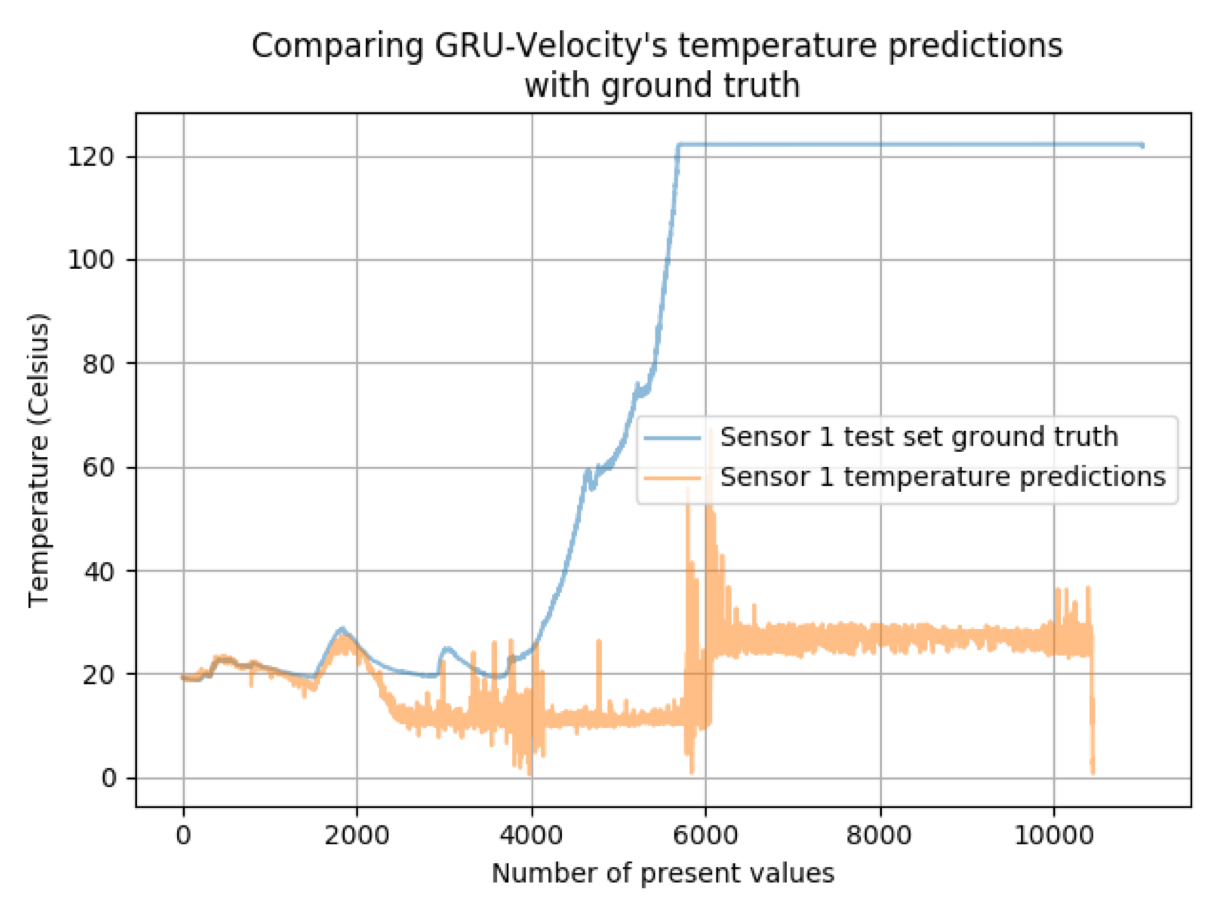}}
    \caption{GRU-Velocity predictions}
    \label{velocityPred}
\end{figure}

\FloatBarrier
Although this added ``momentum" to the predictions as will be seen in the next plot (Figure \ref{velocityPred}), the predictions for all sensors still remain highly correlated.

\FloatBarrier

Notice that the orange curve in Figure \ref{velocityPred} is not directly affected by the other sensor now due to some ``momentum" at around step 2500.
Though the overall predictions don’t improve and the predictions remain correlated.

\section{Future Work}

To overcome the problems in GRU-Simple, the approach suggested by \cite{shukla2019interpolation} looks promising. A ``covariance layer" can be introduced before feeding the inputs to GRU, which ``decorrelates" the input in proportion to the trust in a particular sensor's input. If the value is abnormally high, the trust should be reduced, and vice-versa.
Other possible extensions are:
\begin{itemize}
    \item Multistep forecasting: sequence to sequence models
    \item Imputation: some learned mechanism instead of a forward imputation.
    \item Scalability: To scale to more sensors and larger dataset, batch-wise loading with imputation can be used. There are existing techniques in the Keras library where custom data loaders can be written.
\end{itemize}

While \cite{che2018recurrent} presents the GRU-Decay model which can be broadly applied to datasets with missing values, their assumption that the missing value decays to the mean value may not always hold in datasets. For instance, in stock market prediction, there is no meaningful mean value to decay to. Also, as the authors themselves state, the GRU-Decay model cannot be used in unsupervised settings without prediction labels. Furthermore, all the recurrent neural network models are discrete in nature whereas the multivariate irregularly sampled time series is more suited to a continuous modelling of the data as we are not dealing with just fixed time steps anymore. 
So, a general model which can work in supervised and unsupervised settings without the assumptions of discreteness and decaying to some constant value would address the broadest range of problems in this setting. \cite{DBLP:journals/corr/abs-1907-03907} looks promising.
\nocite{*}
\bibliographystyle{unsrt}  
\bibliography{references}  

\end{document}